\documentclass{svproc}
\usepackage{url}

\usepackage{graphicx}
\usepackage{amsmath}
\usepackage{algorithm}
\usepackage{algorithmic}
\usepackage{xspace}

\begin{document}
\mainmatter              %
\title{Optimal Object Placement using a Virtual Axis}
\titlerunning{Placement using Virtual Axis}  %
\author{Martin Wei{\ss}}
\authorrunning{Martin Wei{\ss}} %
\tocauthor{Martin Wei{\ss}}
\institute{Ostbayerische Technische Hochschule Regensburg\\
Faculty of Computer Science and Mathematics\\ 
Regensburg, Germany\\
\email{martin.weiss@oth-regensburg.de}\\ 
}

\def\frames{{\bbbf}}
\def\SO#1{\mbox{SO}( #1 )}
\def\axis{q}
\def\rotx#1{\hbox{R}_x({#1})}
\def\roty#1{\hbox{R}_y({#1})}
\def\rotz#1{\hbox{R}_z({#1})}
\def\transx#1{\hbox{T}_x({#1})}
\def\transy#1{\hbox{T}_y({#1})}
\def\transz#1{\hbox{T}_z({#1})}
\def\trans#1#2#3{\hbox{Trans}({#1}, #2, #3)}
\def\tcp{\ifmmode \mbox{\sf TCP} \else \mbox{\sf TCP}\xspace\fi}
\def\TCP{\tcp}
\def\tool{\mbox{\sf TOOL}}
\def\atan2#1#2{\mbox{atan2}({#1}, {#2})}
\def\world{{\sf W}}
\def\workspace{{\cal W}}
\def\wcp{\ifmmode \mbox{\sf WCP} \else \mbox{\sf WCP}\xspace\fi}
\def\WCP{\wcp}
\def\equote#1{``#1''}

\renewcommand{\algorithmicrequire}{\textbf{Input:}}
\renewcommand{\algorithmicensure}{\textbf{Output:}}

\maketitle              %

\begin{abstract}
A basic task in the design of a robotic production cell is the relative placement
of robot and workpiece. The fundamental requirement is that the robot can reach all
process positions; only then one can think further optimization. Therefore an algorithm 
that automatically places an object into the workspace
is very desirable. However many iterative optimzation algorithms cannot guarantee that 
all intermediate steps are reachable, resulting in complicated procedures.
We present a novel approach which extends a robot by a virtual prismatic joint
 - which measures the distance to the workspace - such that any TCP frames are reachable. 
This allows higher order nonlinear programming algorithms 
to be used for placement of an object alone as well as the optimal placement under 
some differentiable criterion.
\end{abstract}
\section{Problem Statement}

We consider the following task: A robot should unload a storage box with a chess-board
like structure containing $B_x\times B_y$ identical workpieces at positions $P_{kl}$, 
$k=1, \ldots, B_x$, $l = 1, \ldots, B_y$, counted in the 
coordinate directions of the frame $C\in\frames$ 
(where $\frames$ denotes the set of all frames $\bbbr^3\times\SO 3$) associated with the 
box at distances $D_x$ and $D_y$. Think of test-tubes in medicine or small parts 
in general production. The cell setup is considered fixed, so only the placement of the box 
in the cell can be chosen, e.g. because a new order type.

Usually a pick-and-place operation is programmed at one corner only, the other
position commands are computed from this corner position and the indices and distances. 
However it is difficult for the user to assess whether all positions are reachable because of nonlinearity and
axis limits. Testing the corners is a heuristic that works in many cases but there is
no guarantee, so one has to run time-consuming tests. When the process needs to work
on an object from different sides or with different orientations the situation is even
more complicated. So the user would like to have an algorithm that determines a feasible
object frame $C$ near some initial guess $C_0$, maybe additionally optimizing one of the
many known manipulability measures, see \cite{Merlet}, \cite{Yoshikawa}.

It is easy to check in a program whether a given frame $C$ leads to
reachable positions or not
but it is difficult for a nonlinear optimizer to determine a
direction that leads to a ''{more feasible}'' situation: Feasibility is a binary
decision; the backward transformation 
will usually issue an error only, and abort.

Our idea is to introduce a virtual joint as a {\em slack variable} in 
terms of nonlinear programming (see \cite{Nocedal}) into the optimization
problem that measures the distance of a position from feasibility. This variable 
therefore has an intuitive geometric interpretation.
We can also interpret the virtual axis as a homotopy variable similar to
\cite{Garcia:Zangwill} which gives an ''easy'' solution for large values.
Our slack variable is not generated by the standard procedure replacing an inequality 
constraint $g(x)\leq 0$ by the equality $g(x)+z=0$, with the sign constraing $z\geq 0$. 

Our approach has some similarity to the introduction of virtual axes 
for singularity avoidance in 
\cite{Reiter} or \cite{Leontjevs}. However we do not introduce a rotational 
joint to reduce velocities near singularities but rather use a prismatic joint to
enlargen the mathematical workspace in the optimization process. In combination 
with a smooting operation we can use standard optimization algorithms which require
differentiability of order 1 like all algorithms based on gradient descent, 
or order 2 like Sequential Quadratic Programming (SQP), cf.~\cite{Nocedal}.
Our approach is not related at all to voxelization of the workspace like in 
\cite{Vahrenkamp} and other algorithms aiming at collision free planning.

\section{Virtual Axis Approach}

For ease of exposition we choose a 6R robot resembling the well known Puma 560 
but with more zeros in the parameters. We could extend all formulae 
to similar 6R real industrial robots. We use the DH convention 
$$
	\rotz {\axis_i} \cdot \transz {d_i} \cdot \transx {a_i} \cdot \rotx{\alpha_i} =: 
	\rotz {\axis_i} \cdot B_i =: 
	A_i(\axis_i) 
$$
to get the wrist centre point (\wcp) and tool centre point (\tcp)
\begin{eqnarray*}
   \wcp(\axis) &=& A_1(\axis_1)\cdot  A_2(\axis_2)\cdot  A_3(\axis_3) \cdot 
		A_4(\axis_4) \cdot A_5(\axis_5) \cdot A_6(\axis_6) 
\\
	\tcp(\axis) &=& \wcp(\axis) \cdot \tool
\end{eqnarray*}
expressed relative to the world coordinate system chosen as the axis 1 coordinate 
system. Note that $a_6 = d_6 = 0$ in our case so the DH chain ends in the \wcp as the
essential point for the backward transform.
\begin{figure}
\hfill
  \begin{tabular}{c|c|c|c|c||c}
        $i$ \   & \ $\theta_i$ \ & \ $d_i$ \   &  \  $a_i$ \ & \ $\alpha_i$ \ & \ type\\ 
	\hline
         1   & $q_1$ & 0   & 0  &  $\frac \pi 2$ & R\\
         2   & $q_2$ &  0 &   $l_{23}$ & 0 & R\\
         3   & $q_3$ &  0 & 0 & $-\frac \pi 2$ & R\\
         4   & $q_4$ &  $l_{35}$ & 0 & $\frac \pi 2$ & R\\
         5   & $q_5$ &  0 & 0  & $-\frac \pi 2$ & R\\
         6   & $q_6$ &  0   & 0 & 0 & R
  \end{tabular}
\hfill
\begin{minipage}{60mm}
\includegraphics[width=60mm]{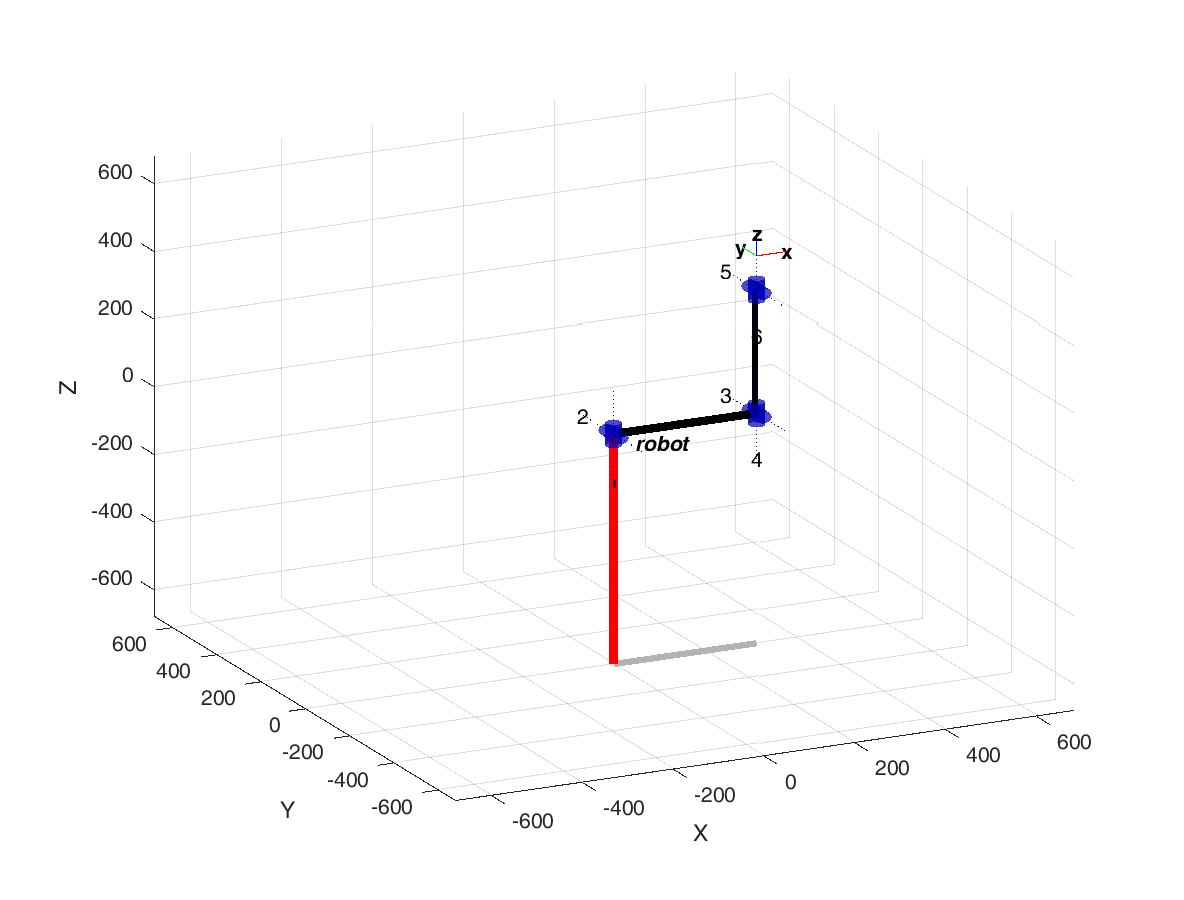}
\end{minipage}
\hfill\ 
  \caption{DH parameters and and reference position 
       $\axis=0$ for original robots}
  \label{fig:DHparameters}
\end{figure}
Figure \ref{fig:DHparameters} 
shows the robot data and the reference position. 
We use $l_{23}=315$, $l_{35} = 365$ and a tool with 
$\transz{t_z}$, $t_z=100$, pointing in direction $z_6$ from the flange.
Lengths are measured in [mm], angles in [rad].
Note that joint 3 is pointing upward for $\axis=0$, so the stretched position 
corresponds to $\axis_3 = -\frac \pi 2$. We assume joint limits
$-\pi \leq \axis_{\min,i} \leq \axis_i \leq \axis_{\max,i} \leq \pi$, $i=1, \ldots 6$. 
As usual, this 
type of robots has up to 8 discrete solutions of the backward transform for 
non-singular positions. We identify these 8 configurations with an integer 
$s\in\{0, \ldots, 7\}$.

Infeasibility of the backward transform for a given frame $F$ and configuration
$s$ may arise from two reasons with different severity: 
First, the \wcp may be to far from the robot such that
the triangle construction for $\axis_3$ fails. There is no remedy in this case. 
Second, even if axis values $\axis$ exist such that $\tcp(\axis) = F$, these might 
violate the joint limits: $\axis_i\not\in [a_i, b_i]$ for some $i$. 
This is no obstacle during the optimization process, only for a solution. 
So the second problem can be fixed by dropping the joint limits and allowing
$\axis_i \in (-\pi, \pi]$, $i=1, \ldots, 6$. 
We describe this difference by a {\em physical} and a {\em mathematical} workspace
$\workspace_P$ and $\workspace_M$ which are the \wcp frames under the two 
joint restrictions under consideration:
\[
	\workspace_P = \wcp\left(\prod_{i=1}^6 [\axis_{\min,i}, \axis_{\max,i}]\right), 
\qquad 
   \workspace_M = \wcp\left((-\pi, \pi]^6\right)
\]
We have 
$\workspace_P\subseteq \workspace_M$ and $\workspace_P\neq \workspace_M$ in general
but $\workspace_M$ is still a bounded set, which is the first problem.

\begin{figure}
\hfill

  \begin{tabular}{c|c|c|c|c||c}
        $i$ \   & \ $\theta_i$ \ & \ $d_i$ \   &  \  $a_i$ \ & \ $\alpha_i$ \ & \ type\\ 
	\hline
         1   & $\tilde q_1$ & 0   & 0  &  $\frac \pi 2$ & R\\
         2   & $\tilde q_2$ &  0 &   $l_{23}$ & 0 & R\\
         3   & $\tilde q_3$ &  0 & 0 & $-\frac \pi 2$ & R\\
         4   & 0 & $v$ & 0 & 0 & P\\
         5   & $\tilde q_4$ &  $l_{35}$ & 0 & $\frac \pi 2$ & R\\
         6   & $\tilde q_5$ &  0 & 0  & $-\frac \pi 2$ & R\\
         7   & $\tilde q_6$ &  0   & 0 & 0 & R
  \end{tabular}
\hfill
\begin{minipage}{60mm}
\includegraphics[width=60mm]{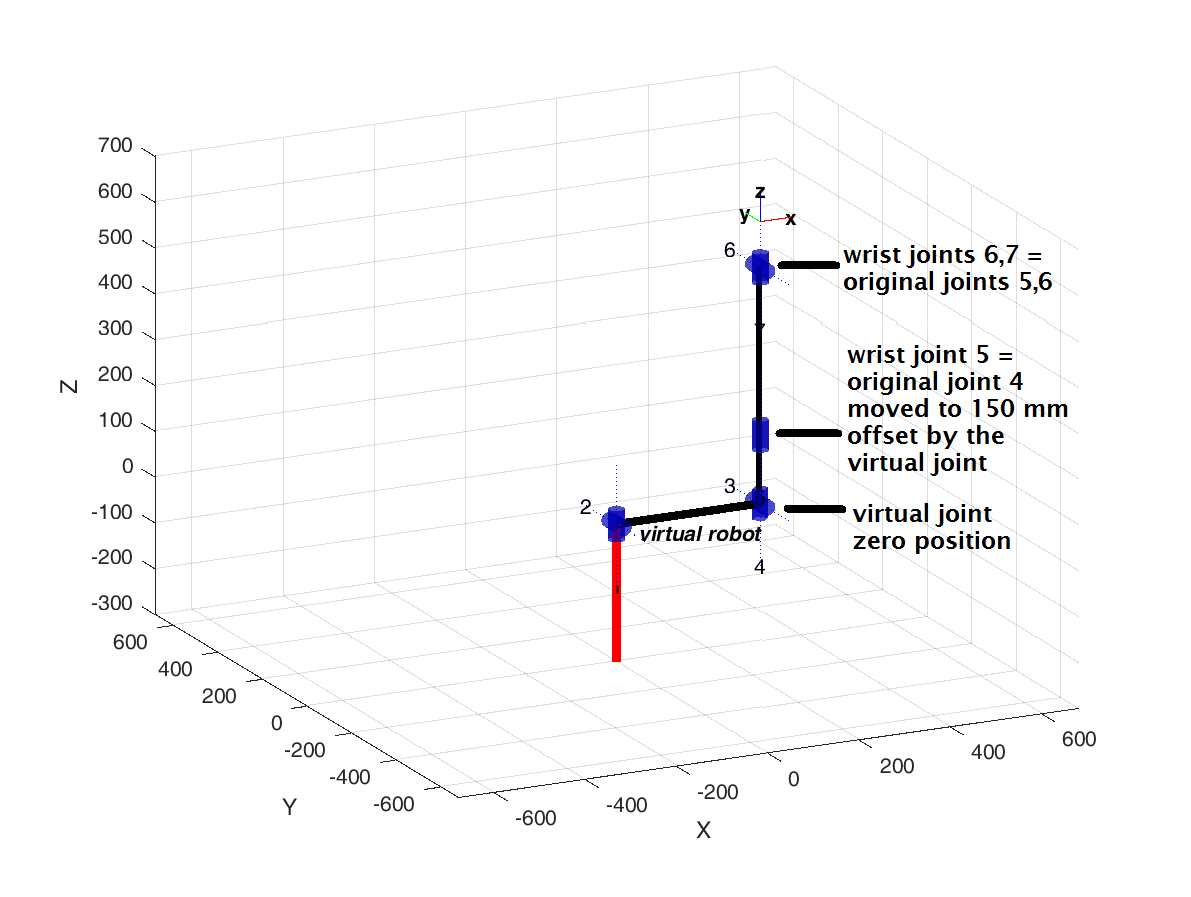}
\end{minipage}
\hfill\ 
  \caption{DH parameters and reference position 
       $\tilde\axis=(0,0,0,150,0,0,0)$ for virtual robot}
  \label{fig:DHparametersVirtual}
\end{figure}

In order to use optimization algorithms which may leave the feasible set $\workspace_M$,
our goal is to define a {\em virtual robot} which has a solution for the backward transform
for any frame $F\in\frames$ and any configuration $s$. 
So we associate to our {\em original robot} a {\em virtual robot} with an 
additional {\em virtual prismatic joint}
between joints 3 and 4, which has no joint limits. Any \wcp in $\bbbr^3$ is reachable then.
The variable of the virtual joint will be denoted $v$, the other
joints keep their  names giving a combined joint variable 
$\tilde \axis = (\axis_1, \axis_2, \axis_3, v, \axis_4, \axis_5, \axis_6)\in\bbbr^7$.
DH parameters of the virtual robot are shown in 
Figure \ref{fig:DHparametersVirtual}.

Sufficient conditions for our approach are stated as two assumptions:

\medskip

{\bf Assumption 1 - Reachability of $\bbbr^3$:} The mapping of the 
original joints 
and the virtual joint to the \wcp position is surjective onto $\bbbr^3$.

\medskip

{\bf Assumption 2 - Reachability of $\SO 3$} Joints 4,5,6 form a central wrist 
parametrizing all of $\SO 3$, i.e. 
the mapping $(-\pi, \pi]^3 \to \SO 3$, 
$(\axis_4, \axis_5, \axis_6)\mapsto 
\rotz {\axis_4} \cdot B_4\cdot\rotz {\axis_5} \cdot B_5\cdot\rotz {\axis_6} $ is surjective.

\medskip

In our case Assumption 1 is satisfied because for any fixed $q_3$ value
the virtual joint generates an infinite line with the \wcp, the second
joint rotates the line through a plane, the first joint rotates the plane through 
all $\bbbr^3$. 
Central wrists satisfying Assumption 2 -- for unbounded joint variables -- are the
most common choice in industry. Assumption 2 also guarantees an 8-solution 
kinematics.  Using the notation $\wcp^v$ to distinguish the 
forward transform of the virtual robot we denote 
\[
	\workspace_V = \wcp^v\left((-\pi, \pi]^3 \times \bbbr \times (-\pi, \pi]^3]\right).
\]
Under our assumptions we get $\workspace_V = \frames$. 
We call such a robot a {\em dextrous robot} because the 
dextrous workspace in $\bbbr^3$ (points reachable with all orientations, see \cite{Craig})
and reachable workspace (points reachable with at least one orientation)
coincide, and are all of $\bbbr^3$.

\section{Backward Transform with Redundancy Resolution}

However we have introduced redundancy in our kinematics so we have to define a backward
transform giving unique results. The virtual robot backward transform sets the virtual 
joint to the smallest absolute value such that a solution exists. 
In our case this is the 
distance between the \wcp position and $\workspace_M$ which is a hollow sphere for our robot
so calculations are simple. Algorithm 1 uses the backward transform of the original robot:
\begin{algorithm}[ht]
\caption{virtual robot backward transform}
\begin{algorithmic}
\REQUIRE $\wcp$ frame $F$ parametrized by $P = [x,y,z]\in\bbbr^3$, $Q\in\SO 3$
\REQUIRE configuration coded as $s\in\{0, \ldots, 7\}$
\ENSURE virtual robot axes $\tilde q\in (-\pi, \pi]^3 \times \bbbr \times (-\pi, \pi]^3]$
with $\wcp(\tilde q)=F$ where $\tilde q$ corresponds to configuration $s$
\IF {solution $q\in (-\pi, \pi]^6$ of original backward transform exists }
  \STATE return essentially this solution as $\tilde q = (q_1, q_2, q_3, 0, q_4, q_5, q_6)$
\ELSE
  \STATE compute $q_1$ using the original backward transform: $q_1 = \atan2{y}{x}$
  \STATE $d:= ||P||_2$, distance of \wcp from robot base 
  \IF {$P$ is outside of the hollow sphere, $d>l_{23}+l_{35}$}
      \STATE $v:= d-l_{23}-l_{35}$
      \STATE put joints 2 and 3 in stretched position $q_2:=\atan2{z}{r}$, $q_3:=-\frac \pi 2$
  \ELSE
      \STATE {$P$ is in the empty interior of the hollow sphere, $d< |l_{23}-l_{35}|$}
	   \STATE $v:= d-l_{23}+l_{35}$
      \STATE put joints 2 and 3 in stretched position $q_2:=\atan2{z}{r}$, $q_3:=+\frac \pi 2$
  \ENDIF 
\ENDIF
\end{algorithmic}
\end{algorithm}
Note that both $l_{23}>l_{35}$ and $l_{23}<l_{35}$ lead to an empty interior, 
but $l_{23}<l_{35}$ results in negative $v$ values.
Also note that for our robot the stretched position for axis 3 is 
$q_3=\pm \frac\pi 2$, not 0 as in most industrial robots. This stretched position is
used for all \WCP positions outside $\workspace_M$.

For robots other than our simple one the computation of $q_1$, as well as 
the definition of the stretched position and the computation of $v$
have to be adapted.

\section{Smoothness properties}

\begin{figure}
\centerline{
\includegraphics[height=35mm]{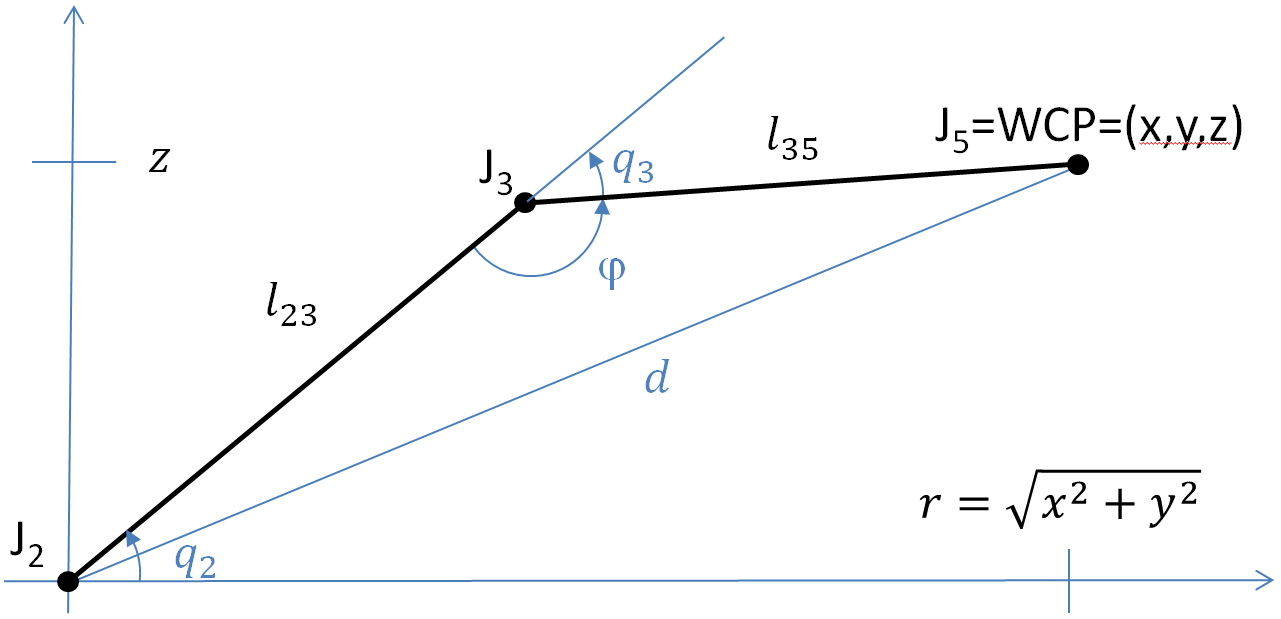}
\includegraphics[height=35mm]{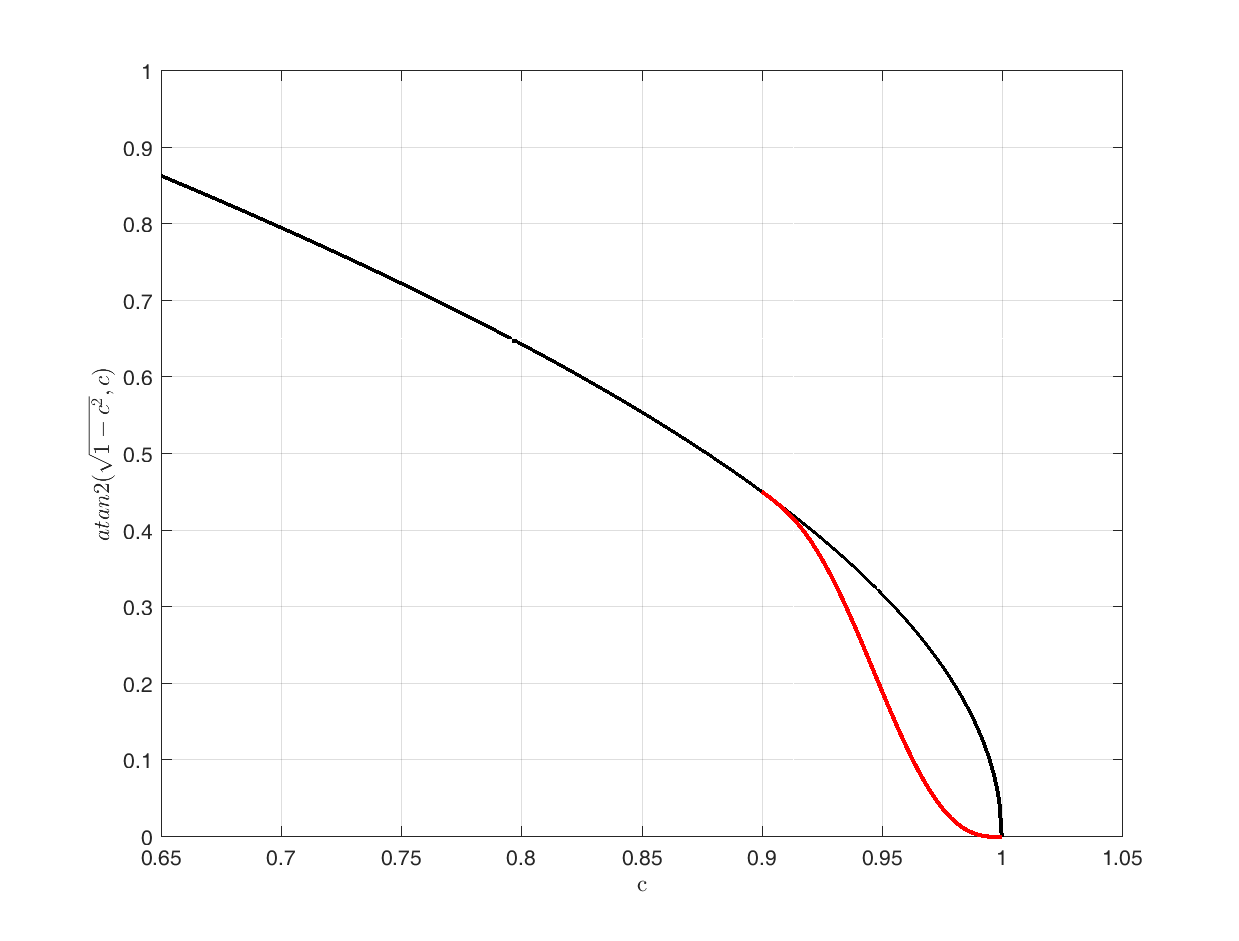}   %
}
\caption{Backward transform and smoothing}
\label{figBackwardTransform}
\end{figure}

The dexterity of the virtual robot makes all $\tcp$ frames feasible. Also the 
joint values $\tilde q$ depend continuously on the \tcp frame, if we keep the 
configuration fixed and avoid singularities or the original robot.
Inside $\workspace_M$ continuity is clear from the usual backward transformation formulae. 
When the 
\wcp reaches the boundary of $\workspace_M$ the triangles for the elbow-up and 
elbow-down configuration degenerate to a line and coincide, hence $q_3 \to -\frac \pi 2$
which is the same value as when approaching from outside. Also $q_2 = \atan2 {z}{r}$ 
depends continuously on the wrist centre point. However the dependence is not 
differentiable: When the wrist centre point enters $\workspace_M$ from outside 
the solution for $q_3$ applies the cosine theorem to get $c=\cos(\varphi)$ near 
$\phi=\pi$. Computing $\varphi(c)=\atan2 {\pm \sqrt{1-c^2}} {c}$ we get an infinite slope at
$c=1$, see 
Figure \ref{figBackwardTransform}. 
This non-differentiability can affect
all other axes.

In order to get $C^2$ behaviour of all angles we modify $\varphi(c)$ on some interval
$[1-\epsilon, 1]$ with a polynomial of degree 5 with $C^2$ transition at $c=1-\epsilon$ and
first and second derivative 0 at $c=1$, see the red graph in 
Figure \ref{figBackwardTransform}, right. This is a basic idea in nonsmooth 
optimization \cite{Bertsekas:kinks}. Of course this distorts the backward transform and
one has to check whether smoothing affects the optimal solution but the hope is that
the optimum is sufficiently inside the workspace, and smoothing is only a temporary 
help for the optimization algorithm.

\begin{figure}
\centerline{
\includegraphics[height=40mm]{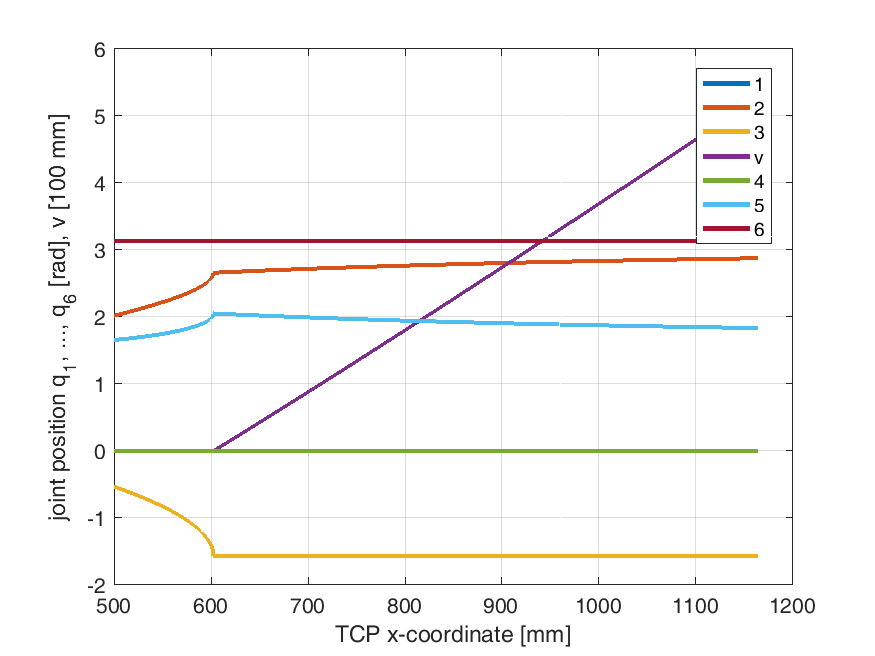}
\includegraphics[height=40mm]{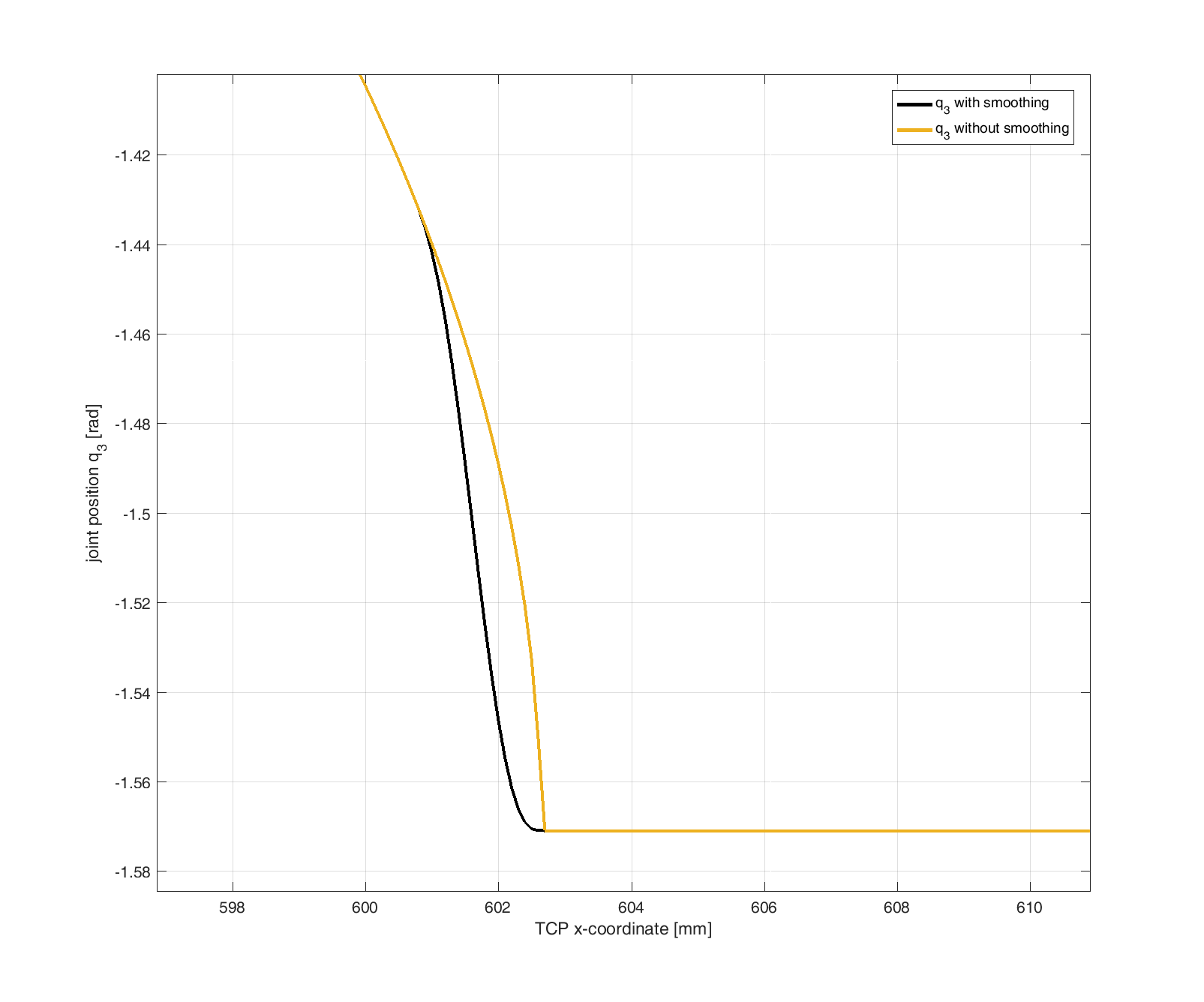}
}
\caption{Joint values for motion through workspace boundary}
\label{fig:MoveRobotAlongX}
\end{figure}

Figure \ref{fig:MoveRobotAlongX} shows this behaviour for a motion of the virtual robot
\tcp from $L_0$ along the $x$ axis to $L_1$ in height $z=215$ with constant 
orientation $Q$ with 
$$
L_0 = \begin{bmatrix} 500\\ 0 \\ 215 \end{bmatrix}, 
\qquad
L_1= \begin{bmatrix} 1300\\ 0 \\ 215 \end{bmatrix},
\qquad
Q = \begin{bmatrix} -1 & 0 & 0 \\ 0 & 1 & 0 \\ 0 & 0 & -1 \end{bmatrix}
$$
in the elbow-up configuration such that the tool is pointing downwards. 
The \wcp\xspace crosses the boundary $\partial \workspace_M$ at
$$
\partial x = \sqrt{(l_{23} + l_{35})^2 - (215+t_z)^2} = \sqrt{680^2-315^2} \approx 602.6.
$$
Axes $q_1, q_4, q_6$ do not move. All
joints are continuous indeed with infinite slope of $q_2, q_3, q_5$ at $\partial x$.
The left picture shows the axis values without smoothing, the right 
a zoom into the $q_3$ behaviour without and with smoothing.
The virtual axis remains at $v=0$ in $\workspace_M$, then $v$ grows linearly. 
So the virtual joint $v(x)$ is also non-differentiable at $\partial x$ but has
essentially the same behaviour like $x\mapsto \max\{0,x\}$ at $x=0$. Forming the
square $x\mapsto (\max\{0,x\})^2$ creates a $C^1$ function, we will use this trick 
in our objective function.

\section{Formulation of the Optimization Problem}

We assume that poses must be reached with the same configuration; this is quite
usual for Cartesian task. We parametrize the corner frame as 
$C=\trans x y z \cdot \rotz \alpha \cdot\roty \beta \cdot\rotx \gamma$. These parameters
or a subset thereof constitute our optimization variables.
Denoting $\tilde q^{(k,l)} = \tilde q^{(k,l)(C)} = 
\tilde \axis^{(k,l)}(x,y,z,\alpha,\beta, \gamma)$ the joint values obtained
by the virtual robot backward transform for grid position
$(k,l)$ in the box, 
$\axis_i^{(k,l)}$ and $v^{(k,l)}$ the original and virtual joints, we may optimize
\begin{eqnarray*}
	\min_{x,y,z,\alpha,\beta, \gamma} & &
	\sum_{k=1}^{B_x}
	\sum_{l=1}^{B_y} (v^{(k,l)})^2
\\
\text{under} && \axis_{\min,i} \leq \axis_i^{(k,l)} \leq \axis_{\max,i}\\
& &  i=1, \ldots, 6, 
k = 1, \ldots, B_x, l = 1, \ldots, B_x
\end{eqnarray*}
We can also add constraints on the frame parameters $(x,y,z,\alpha,\beta, \gamma)$.
The square in the objective function makes the objective function $C^1$.
To make advantage of this however we must use the $C^2$
smoothing of the backward transform at the workspace boundary so that the constraints 
are $C^2$ as well. With $|v^{(k,l)}|^3$ we could even obtain a $C^3$ objective function.

Adding some manipulability criterion from \cite{Yoshikawa} or \cite{Merlet} to the 
objective function can optimize feasibility and manipulability in combination. 
However one has to use appropriate weighting because in extreme cases the optimizer 
might tolerate some infeasible points in exchange for high manipulability at other 
points.

\section{Numerical Results}

\begin{figure}
\centerline{
\includegraphics[height=40mm]{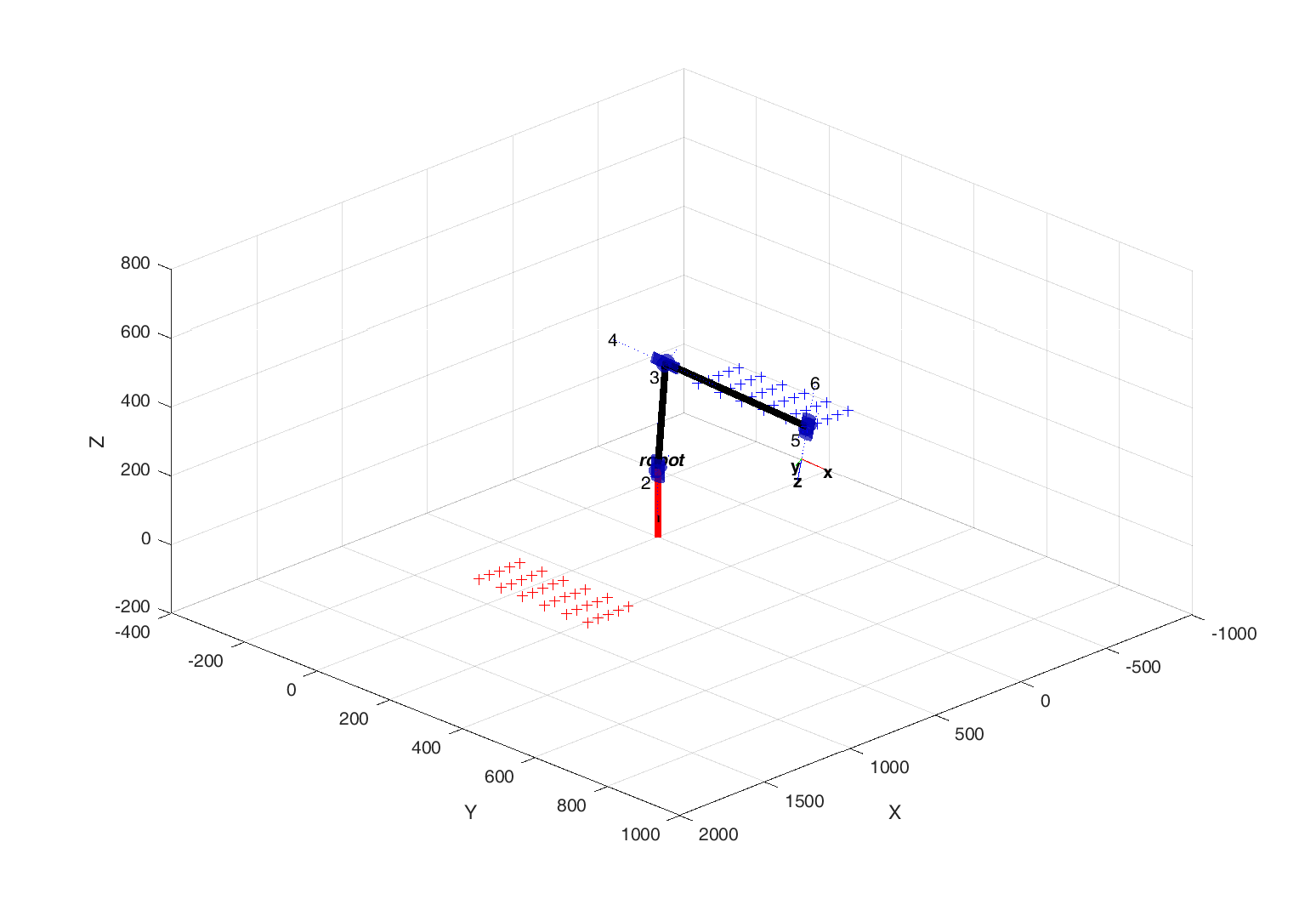}
\includegraphics[height=40mm]{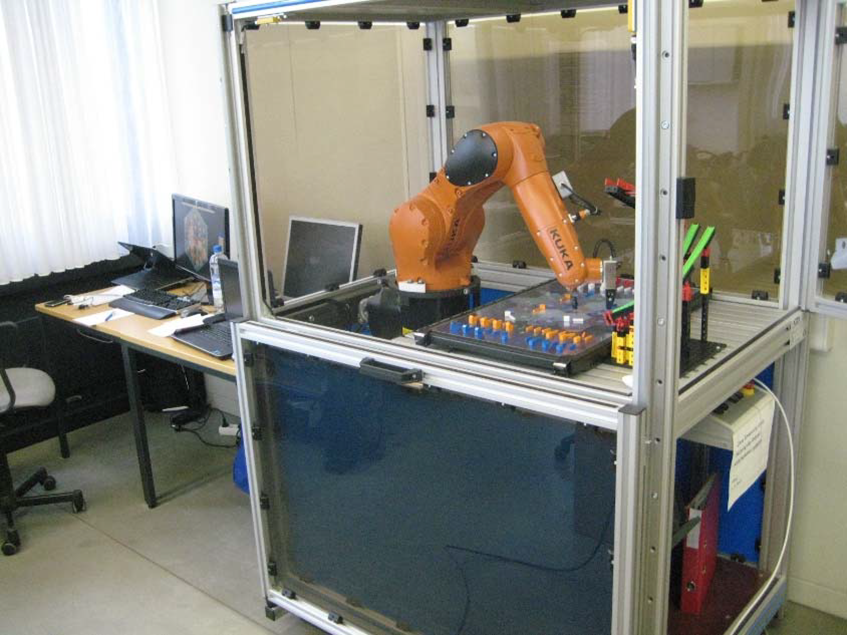}
}
\caption{Optimization results and demonstrator setup}
\label{fig:optimizeGridPosition}
\end{figure}

We have tested our optimization procedure with the solvers implemented in the MATLAB 
{\tt fmincon} command. We obtained optimal solutions both with the default 
interior point algorithm and the SQP algorithms. However, in many cases the SQP algorithm
required only 10 iterations, about half the iterations of the interior point algorithm. 
Computation time was below 10 sec on a standard laptop with $B_x\cdot B_y = 5\cdot 6 = 30$
grid point in the box. 
Figure \ref{fig:optimizeGridPosition} 
shows some typical run where a box is drawn from far
outside the workspace (red) to the interior (blue). The results were almost independent
from differentiability properties. The interpretation is: Even when crossing the critical 
workspace boundary the majority of grid points is away from the vertical slope, dominating
the numerical derivatives. When sufficiently inside the workspace the algorithm hardly 
ever enters the infeasible region again.

We have also tested the algorithm with success for a more complicated geometry: 
The robot makes the moves for the popular board game 
''Settlers of Catan'' with real tokens on a hexagonal structure on 
a playfield simulated on a screen, see \cite{KUKA:StudentsWin}. 
However some manual intervention is necessary as the robot cannot reach all positions
with the same configuration.

\section{Conclusions}

The approach presented in this contribution opens a way to non-interior-point 
optimization algorithms for the most common class of industrial 6R robots.
Obviously the approach can also be used 
for the placement of a robot in a fixed work cell; we only need to consider the 
robot's base coordinate system as the variable. Furthermore it seems promising
to use the idea for the optimization of redundant tasks like in \cite{Leger:Angeles}
or redundant robots and to compare the results.

New algorithms for non-differentiable problems 
\cite{Curtis:Overton} should be investigated in comparision to higher order approaches, 
including estimates for the convergence of the solution of the smoothed problem to the 
original problem's solution.


\begin{thebibliography}{6}
%

\bibitem{Bertsekas:kinks}
Bertsekas, D.P.: Nondifferentiable Optimization via Approximation. 
Math. Programming Study 3, pp. 1-25 (1975).


%
%
%
%
%

%
%

\bibitem{Craig}
 Craig, J.: Introduction to Robotics: Mechanics and Control.
 Addison Wesley, (1986)

\bibitem{Curtis:Overton}
Curtis, F.E., Overton,  M.L.:
A Sequential Quadratic Programming Algorithm for Nonconvex, 
Nonsmooth Constrained Optimization.
SIAM J. Optimization 22, pp. 474-500 (2012)

\bibitem{Garcia:Zangwill}
Garcia, C., Zangwill, W.: Pathways to solutions, fixed points and equilibria.
Prentice-Hall (1981)

%
%
%
 
\bibitem{KUKA:StudentsWin}
KUKA Robot Group: Robots Play Board Games - Students Win Big, 
\url{https://www.youtube.com/watch?v=oCQPWv_ky2c}, (2016)

\bibitem{Leger:Angeles}
L{\'e}ger, J., Angeles, J.:
Off-line programming of six-axis robots for optimum
five-dimensional tasks.
Mechanism and Machine Theory 100, 155–-169  (2016)

\bibitem{Leontjevs}
Leontjevs, V., Flores, F.G., Lopes, J., Kecskemethy, A.:
Singularity Avoidance by Virtual Redundant Axis and its Application to 
Large Base Motion Compensation of Serial Robots
In: Proceedings of the RAAD 2012
21st International Workshop on Robotics in Alpe-Adria-Danube Region.
Naples (2012)

%
%

\bibitem{Merlet}
Merlet, J.P.: Jacobian, manipulability, condition number, and accuracy of parallel robots. J.
Mech. Des. 128(1), 199-–206 (2006)

\bibitem{Nocedal}
 Nocedal, J., Wright, S.J.:
 Numerical Optimization.
 Springer, New York (2006)

\bibitem{Reiter}
Reiter, A.: Ein Beitrag zur Singularit\"atsvermeidung bei
Industrierobotern durch Einf\"uhrung virtueller Achsen.
Master Thesis, Johannes Kepler University Linz (2015)

\bibitem{Vahrenkamp}
Vahrenkamp, N., Asfour, T., Dillmann, R.:
Robot Placement based on Reachability Inversion.
2013 IEEE International Conference on Robotics and Automation (2013)

\bibitem{Yoshikawa}
Yoshikawa, T.: Manipulability of Robotic Mechanisms  
The International Journal of Robotics Research 4(2), pp. 3-9 (1985).


%
%
%

%
%
%
%
%
%
%
%
%
%
%
%
%
%
%
%
%
%
%
%
%
%
%
%
%
%
%


\end{thebibliography}
\end{document}